\documentclass{llncs}

\usepackage{graphicx}

\begin{document}

\title{On how percolation threshold affects PSO performance}

\author{Blanca Cases, Alicia D'Anjou, Abdelmalik Moujahid}

\institute{Computational Intelligence Group of the University of the Basque Country. \\
P. Manuel de Lardiz{\'a}bal n. 1, 20018 San Sebasti{\'a}n, Spain
}

\maketitle

\begin{abstract}
Statistical evidence of the influence of neighborhood topology on the performance of particle swarm optimization (PSO) algorithms has been shown in many works. However, little has been done about the implications could have the percolation threshold in determining the topology of this neighborhood. This work addresses this problem for individuals that, like robots,
are able to sense in a limited neighborhood around them. Based on the concept of percolation
threshold, and more precisely, the disk percolation model in 2D, we show that better results are obtained for low
values of radius, when individuals occasionally ask others their best
visited positions, with the consequent decrease of computational complexity. On the other hand, since percolation threshold is a universal measure, it could have a great interest to compare the performance of different hybrid PSO algorithms.
\end{abstract}

\keywords{Particle Swarm Optimization parameters, Percolation theory, Hybrid PSO.}

\section{Introduction}
Percolation theory appears in very different random structures, including spread of diseases, fire propagation in forests, phase transition in solids, diffusion in disordered media, etc. There are number of good reviews of the percolation theory \cite{Sahimi_1994,Stauffer_1992}. This theory is particularly well adapted to describe global physical properties, such as the connectivity and
conductivity behaviour of geometrically complex systems. This work analyzes the role of percolation threshold, a universal measure of connectivity in graphs, to analyze the convergence of swarms algorithms as a function of the expected number of neighbors at initial step. And therefore, to develop a framework to compare the performance of hybrid PSO algorithms. The analysis of the performance of different hybrid PSO algorithms has been addressed in different works \cite{DBLP:journals/ijon/CorchadoGW12,García_Fernández_Luengo_Herrera_2010,DBLP:journals/isci/CorchadoAC10}. However, in this work, we focus on the concept of percolation threshold as a useful tool to define the parameter space for which the performance of the basic PSO algorithm is enhanced.

The basic model of PSO by Eberhart, Kennedy and Shi \cite{Kennedy1997303,Eberhart200181}
is an stochastic optimization method for \emph{D}-dimensional functions.
Given a population of $P$ individuals i, $1\leq i\leq P$, uniformly
distributed on an \emph{D}-dimensional hypercube of size $S$ in points
$x_{i}=\left(x_{i1},\ldots,x_{id},\ldots,x_{iD}\right)$, each individual
moves to next position according to a velocity vector $v_{i}=\left(v_{i1},\ldots,v_{id},\ldots,v_{iD}\right)$
both calculated according to (\ref{eq:PSO}):
\begin{equation}
{\small
\begin{array}{ll}
v_{id}  =  w*v_{id}+c_{1}*rand\left(1.0\right)*\left(p_{id}-x_{id}\right)+c_{2}*Rand\left(1.0\right)*\left(p_{g_{i}d}-x_{id}\right)\\
x_{id}  =  x_{id}+v_{i}
\end{array}
\label{eq:PSO}
}
\end{equation}
where values $rand(1.0)$ and $Rand(1.0)$ are drawn at random according
to a uniform distribution in the interval $[0,1]$. Velocity $v_{i}\left(t+1\right)$
is a randomized linear combination of three forces: ($i$) inertia, weighted
by parameter $w\in[0,1]$, ($ii$) the personal attraction $c_{1}$ to the
position $p_{i}$ where the minimum value of a function $f\left(p_{i}\right)$
was found, and ($iii$) the social attraction to the position $p_{g_{i}}$
of the best goal remembered by the neighbors of individual $i$ in
a given topology.

In many works, the neighborhood of each individual
includes the whole population and the best goal $p_{g}=p_{g_{i}}$
is visible for all the agents in a blackboard architecture. In these cases, the gain
in efficiency runs in parallel to the lose of autonomy
of agents in the model. For realistic purposes, as in swarm robotics,
the notion of Euclidean neighborhood is a building block since sensors
and effectors have physical limits.
Some authors \cite{Toscano-Pulido2011179,Zhang2010661} have carried out
social network statistical studies to address the effects of neighborhood topology in PSO
algorithms, concluding that the topology used conditions strongly the convergence of these algorithms.

In this work, based on the so called continuum
or disk percolation model \cite{bollobs_percolation_2006}, we investigate the effect
of the percolation threshold according to the topology
of geometric graphs.
A geometric graph is a 2D representation of a graph: given a neighborhood
radius \emph{R} and a square board of size S, let $\{x_{1},\ldots,x_{P}\}$
be the initial positions of individuals composing the swarm selected
at random following a uniform distribution. The geometric graph is
built connecting each node to all neighboring nodes in radius R. Percolation
disk theory allows the calculus of the radius $R$, that ensures a
mean of $a$ neighbors per node (called the area or degree) with parameters of population of $P$ individuals and world size $S$, according to  (\ref{eq:area}).

\begin{equation}
R  = \sqrt{\frac{aS^{2}}{\pi P}}
\label{eq:area}
\end{equation}

We study experimentally the convergence of PSO algorithms
focusing on the concept of percolation. The percolation concept comes
from chemistry and refers to the way a liquid flows through a porous
medium, like a large piece of pumice stone and sets what is the probability
that the center of the stone is wet when we immerse it in water. Other
example is making coffee: if ground coffee is fine grained, under
a determined threshold for the diameter of grain, water stagnates,
while over this threshold water falls through coffee. In terms of
random graphs, percolation is identified to the phenomenon of passing
from numerous small connected components to a connected one as a function
of the number of expected neighbors $a$. The critical value of $a$
receives the name of percolation threshold $a_{c}$.

\begin{figure}
\begin{centering}
\begin{tabular}{ccc}
a)\includegraphics[width=0.27\textwidth]{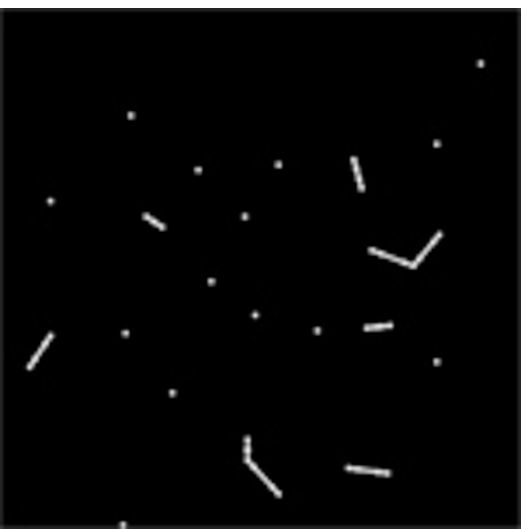} & b)\includegraphics[width=0.27\textwidth]{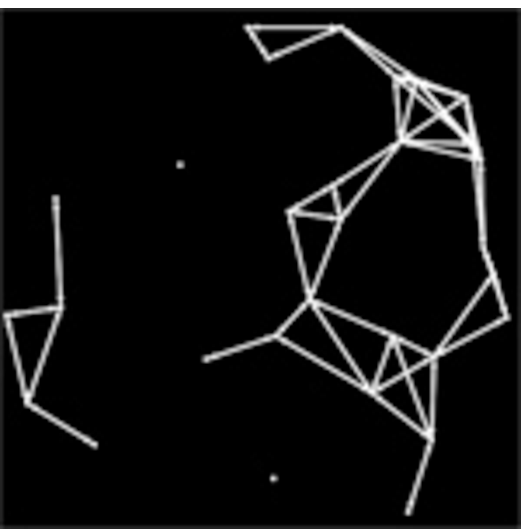} & c)\includegraphics[width=0.27\textwidth]{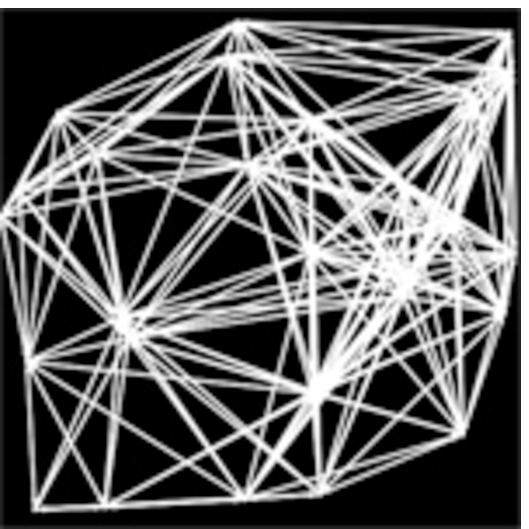}\tabularnewline
\end{tabular}
\par\end{centering} \vspace{0.1cm}

\caption{Random geometric graphs with parameters a) $a=0.8$, $R=18.426$,
b) $a=4.512$, $R=43.760$, c) $a=29$, $R=110.941$}

\end{figure}

Percolation is a complex phenomenon which suggests that agents in
a swarm only need a minimum of social information to perform the optimization
task, being redundancy a source of futile complexity. It has been
proved \cite{bollobs_percolation_2006} with
confidence 99.99\% the following bound for the critical percolation
threshold $a_{c}$ and the corresponding critical value $R_{c}$,
obtained as the average degree over the squares of any size $S$:

\begin{equation}
4.508\leq a_{c}\simeq4.512\leq4.515\label{eq:a_c}
\end{equation}

\begin{equation}
43.741\leq R_{c}\simeq43.758\leq43.775\label{eq:R_c}
\end{equation}

We will show experimentally that neighborhoods of critical radius
$R_{c}$ are enough to reach goals in a minimum number of steps. The
following definitions and parameters were set in order to compare
results with those of \cite{Clerc200258,Eberhart200084,Trelea2003317}.
\begin{itemize}
\item Neighbors are determined by the euclidean distance $d$: let $x_{i}$
be the position of individual $i$, $N_{R}\left(x_{i}\right)=\left\{ x_{i_{1}},\ldots,x_{i_{n_{i}}}\right\} $
is the set of neighboring individuals in radius $R$, that is $d\left(x_{i},x_{i_{j}}\right)\leq R,\:1\leq j\leq n_{i}$
.
\item The neighbors best value of function $f$ known by agent $i$ in its
neighborhood is defined as: $p_{g_{i}}=min{}_{f}\left\{ p_{i_{j}}:\: x_{i_{j}}\in N_{R}\left(x_{i}\right)\right\}$,
that is, the position $p_{i_{j}}$ remembered by neighbors (at current
position $x_{i_{j}}$) minimizing goal function $f$. The social best
of $x_{i}$ is defined as the mean neighbors best value at final step,
$\frac{\sum p_{g_{i}}}{P}$ .

\item Experimental parameters are exactly those given in \cite{Eberhart200084}:
a swarm of $P=30$ individuals and square side $S=200$ and center
(0,0). Velocity module $\left|v\right|=\sqrt{v_{1}^{2}+v_{2}^{2}}$
is limited by $x_{max}=v_{max}=100$, meaning that individuals move
a step forward in the direction of vector $v$ with step length
$s(v)=v_{max}$ if $ \left|v\right|>v_{max}$ and $s(v)=\left|v\right|$  otherwise.
Clerc's \cite{Clerc200258} optimal parameters $c_{1}=c_{2}=1.49445$
and constant constriction factor $w=0.729$ where used.

\item Benchmark functions are those used in \cite{Eberhart200084,Trelea2003317}.
The model has been implemented in Netlogo
4.1, running in concurrent mode (individuals are updated once the
run step ends) with a float  precision of 16 decimals. An experiment
consists in 100 runs for each function, 20 for each instance of parameter
R.
\item We compare results for Clerc's parameters set \cite{Clerc200258,Eberhart200084}
$w=0.729,\; c_{1}=c_{2}=1.49445$ and Trelea's parameters $w=0.6,\; c_{1}=c_{2}=0.7$.
Since disk percolation threshold is a 2D measure, the dimension will
be $D=2$ for the scope of this work. The domain of the benchmark
functions will be $x_{max}=100=v_{max}$, as well as the goal, 0.01
for all functions. In this way results concerning percolation threshold
can be easily compared.
\end{itemize}

\section{Analysis of goals as a function of the expected degree at initial
step.\label{sec:Analysis-of-goals}}

Applying equation (\ref{eq:area}), we explore the success of PSO
algorithm as a function of radius $R(a)$ according to a longitudinal
study of the mean degree expected at initial step.
\begin{itemize}

\item Dataset A: $0.000\leq a\leq0.9$, meaning that only a fraction of
the individuals has neighbors: from 0.0 to 0.9 with step 0.1. Table
\ref{table:R(a)} shows the corresponding neighborhood radius $R(a)$.
Each function is run 20 times, so for each $R\left(a\right)$ 100
runs were completed.

\item Dataset B: $1\leq a\leq30$, from the minimum (1 neighbor) to the
maximum (29 neighbors) plus one, 30 neighbors with step 1. Neighborhood
radius $R$ varies in the interval $20.601\leq R\leq112.838$.
\item Best global value registers at each run step the best position remembered
by any agent, that is $best-global-val\left(t\right)=min\left\{ f\left(p_{i}\left(t\right)\right):\;1\leq i\leq P\right\} $,
updated every time that an individual finds a smaller value $x_{i}\left(t+1\right)<best-global-val\left(t\right)$.
\item Mean best neighbors value is the average social information over all
the individuals, $best-neighbors-val_{i}\left(t\right)=min\left\{ f\left(p_{j}\left(t\right)\right):\;\right\} $,
updated every time that an individual finds a smaller value $x_{i}\left(t+1\right)<best-global-val\left(t\right)$.

\end{itemize}
Varying the benchmark function and the radius of neighborhood as parameters,
each pair 20 runs, with goal precision 0.01 and a maximum number of
iterations $P^{2}=900$, results in table \ref{Flo:goals-best-all}.
Even including the case of neighborhood radius $R=0$, the performance
of the PSO algorithm does better for graphs with initial mean degree
$0\leq a\leq0.9$: with the only exception of $f_{1}$ the mean best
global values are successful reaching the goal $0.01$. As was expected,
the objective function has to do in the performance of the algorithm:
it is harder to optimize function $f_{1}$, but never ceases to amaze
that respecting to the social force, less neighbors means better performance
of PSO. In a further section we will analyze accurately the performance
of the algorithm for values $a=0$ and $0<a\leq0.9$. The mean iterations
(over a maximum of $P^{2}=900$) averaged over all runs is similar
for datasets A and B.

{\footnotesize }
\begin{table}
\begin{centering}
{\footnotesize
\[
\begin{array}{c|c|c|c|c|c|c|c|c|c|c}
a & 0.000 & 0.100 & 0.200 & 0.300 & 0.400 & 0.500 & 0.600 & 0.700 & 0.800 & 0.900\\
\hline R & 0.000 & 6.515 & 9.213 & 11.284 & 13.029 & 14.567 & 15.958 & 17.236 & 18.426 & 19.544
\end{array}
\]
\caption{Values of neighborhood's radius $R(a)$ as a function of expected
degree $a$.}
\label{table:R(a)}}
\end{centering}
\end{table}

\begin{table}
\begin{centering}
{\scriptsize }%
\begin{tabular}{|c||c||c||c|c||c|c||c|c|}
\hline
{\scriptsize Function} & \multicolumn{2}{c||}{{\scriptsize Goals}} & \multicolumn{2}{c||}{{\scriptsize Best global}} & \multicolumn{4}{c|}{{\scriptsize Steps}}\tabularnewline
\cline{2-9}
 & {\scriptsize A} & {\scriptsize B} & {\scriptsize A} & {\scriptsize B} & {\scriptsize A goal} & {\scriptsize B goal} & {\scriptsize A all} & {\scriptsize B all}\tabularnewline
\hline
\hline
{\scriptsize Spherical $f_{0}$} & {\scriptsize 95.50\%} & {\scriptsize 100.00\%} & {\scriptsize 0.01} & {\scriptsize 0.00} & {\scriptsize 92 } & {\scriptsize 56 } & {\scriptsize 129 } & {\scriptsize 56 }\tabularnewline
\hline
{\scriptsize Rosenbrock $f_{1}$ } & {\scriptsize 52.00\%} & {\scriptsize 9.00\%} & {\scriptsize 0.09} & {\scriptsize 0.06} & {\scriptsize 588} & {\scriptsize 370} & {\scriptsize 738} & {\scriptsize 852}\tabularnewline
\hline
{\scriptsize Rastrigin $f_{2}$} & {\scriptsize 95.00\%} & {\scriptsize 100.00\%} & {\scriptsize 0.01} & {\scriptsize 0.00} & {\scriptsize 82} & {\scriptsize 58} & {\scriptsize 123} & {\scriptsize 58}\tabularnewline
\hline
{\scriptsize Griewank $f_{3}$ } & {\scriptsize 100.00\%} & {\scriptsize 100.00\%} & {\scriptsize 0.00} & {\scriptsize 0.00} & {\scriptsize 2} & {\scriptsize 2} & {\scriptsize 2} & {\scriptsize 2}\tabularnewline
\hline
{\scriptsize Schaffer $f_{6}$ } & {\scriptsize 100.00\%} & {\scriptsize 100.00\%} & {\scriptsize 0.00} & {\scriptsize 0.00} & {\scriptsize 5} & {\scriptsize 4} & {\scriptsize 5} & {\scriptsize 4}\tabularnewline
\hline
{\scriptsize Total} & {\scriptsize 88.50\%} & {\scriptsize 81.80\%} & {\scriptsize 0.02} & {\scriptsize 0.02} & {\scriptsize 108} & {\scriptsize 38} & {\scriptsize 199} & {\scriptsize 195}\tabularnewline
\hline
\end{tabular}\label{Flo:goals-best-all}\caption{Percentage of goals, average best global value and iterations over
all runs for datasets A, $0\leq a\leq0.9$, $0\leq R\leq19.544$ and
B, $1\leq a\leq30$, $20.601\leq R\leq112.838$..}
\end{centering} \vspace{0.1cm}
\end{table}

\begin{figure}
\label{Flo:mean-global-all}\includegraphics[width=1\textwidth]{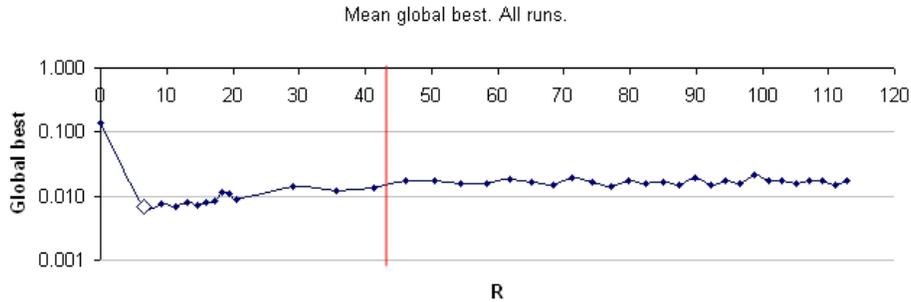}\caption{Mean global best as a function of neighborhood radius R shows that
PSO behaves better under critical $R_{c}\simeq43,758$. Non successful
runs are included. A minimum is obtained at R=6.515 growing until
$R_{c}$ is reached.}
\end{figure}

\begin{figure}
\label{Flo:succesful-runs}\includegraphics[width=1\textwidth]{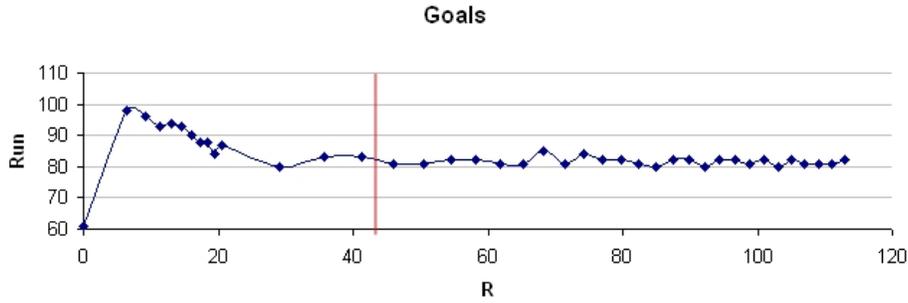}\caption{Number of goals reached as neighborhood R increases.}
\end{figure}

Figure \ref{Flo:mean-global-all} shows mean values of the best global as a function of neighborhood radius $R\left(a\right)$ calculated
according to  (\ref{eq:area}). As it can be appreciated, the best
result corresponds to a disk radius $R\left(a\right)=6.515$ where the
degree $a=0.100$ is the first one being explored. This means that
initial configurations, where 10\% of individuals have a neighbor,
give the best performance. Note that $a<\log\left(\log\left(P\right)\right)=0.169$
since $P=30$ and hence each run has computational complexity at most
of the order of $P\log\left(\log\left(P\right)\right)$. Over the
critical radius $R_{c}$ the mean global best stops increasing to become
constant.
As one might expect, Fig. \ref{Flo:succesful-runs} shows that the
only memory of the best personal value gives comparatively a poor
score. Even so, this score is over the 60\%. The importance
of social information is crucial: surprisingly the smallest number
of initial neighbors gives the majority of the goals. Near of the
percolation threshold $R_{c}\simeq43,758$ the score starts remaining
constant.
Finally, the mean iterations needed for success, with a limit of $P^{2}=900$,
reach the minimum at $R=6.515$. From this value, the mean over all
the runs slightly increases becoming constant, near to 195 steps,
for values upper to $R_{c}\simeq43.758$. On the other hand, the average
over successful runs decreases until the threshold $R_{c}$ is passed
reaching a constant value near in mean to 36 steps.

\begin{figure}
\includegraphics[width=1\textwidth]{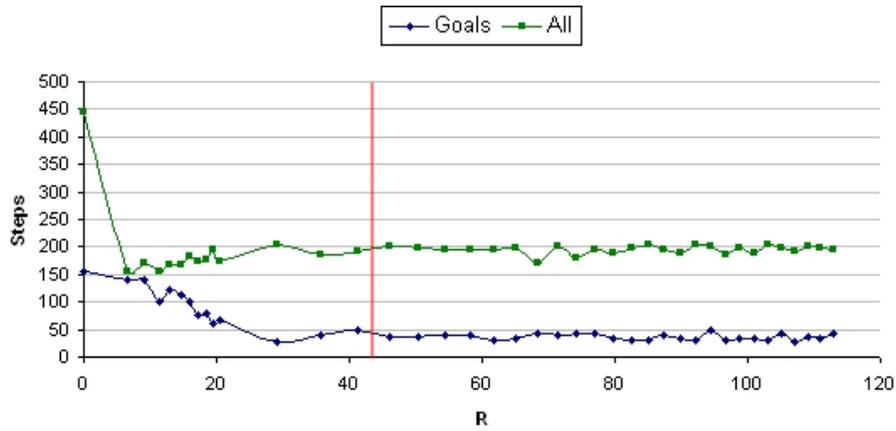}\label{Flo:iterations}\caption{Mean number of iterations. Square marks represent the mean over all
runs and diamonds over successful runs. }
\end{figure}

\begin{figure}
\includegraphics[width=1\textwidth]{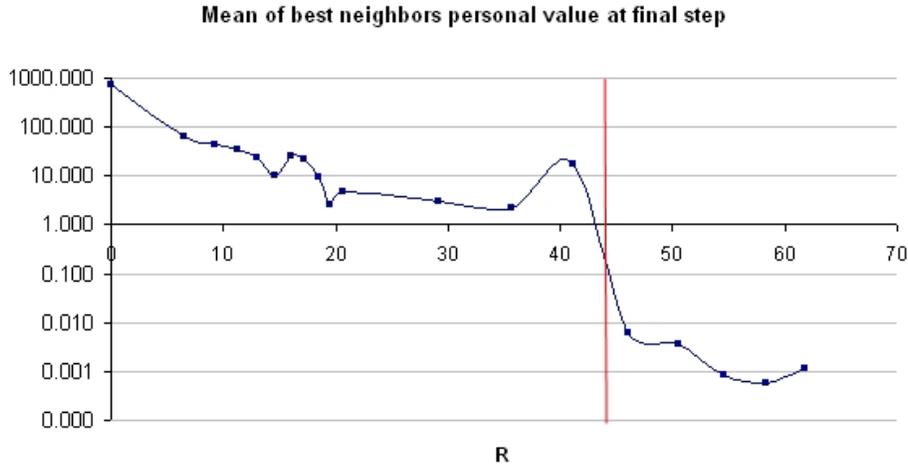}\label{Mean-neighbors-val}\caption{The mean of social best at final step.}
\end{figure}

The mean value of social best, that is the average value of the best
personal value of neighbors, was measured at final step. The mean
over all runs is presented in Fig. \ref{Mean-neighbors-val}. Beyond
value $R=61.804$ the mean reaches value 0 rounded to 3 decimals.
This occurs because some individuals fly outside the limits of the
board having 0 neighbors. This curve, as will be shown in next section,
seems to be very sensible to the variations of $R$, decreasing drastically
at critical radius $R_{c}$.

\section{Exploring the disk percolation threshold\label{sec:Exploring-the-disk}}

Values of $R$ around critical radius $R_{c}$ were finely explored varying
degree $a$ along the interval $[4.37,4.515]$ just before the critical
threshold $a_{c}=4.512$ with step 0.001. Hence $R(a)$ varies in
the interval $[43.066,43.775]$ in intervals of 0.01. Each instance
of parameter $R$ was run 100 times, 20 for each function $f_{i}$.
Table \ref{Flo:goals-best-Rc} shows the means of goals, the best
global value reached and final step. Comparing to table \ref{Flo:goals-best-all},
the total of successful runs equals the goals of dataset A, 88.5\%,
as well as the mean best global value, 0.02.
The difference with dataset A is in the number of steps: at critical
radius $R_{c}$ the performance of PSO algorithm decreases to mean
of 42 iterations in successful runs, near to the 38 steps reached
in dataset B. Mean steps, 197, are just in the middle of the 199 obtained
for dataset A and the 195 of dataset B. As a conclusion, the performance of PSO algorithm with $a_{c}$ between 4 and 5 neighbors is the same as with large values, even when the neighborhood comprises all agents.

\begin{table}
\begin{centering}
\begin{tabular}{|c||c||c||c|c|}
\hline
 & \multicolumn{1}{c||}{{\scriptsize Goals reached.}} & {\scriptsize Best global all.} & {\scriptsize Steps Goals} & {\scriptsize Steps all}\tabularnewline
\hline
\hline
{\scriptsize Spherical $f_{0}$} & {\scriptsize 95.50\%} & {\scriptsize 0.01} & {\scriptsize 57 } & {\scriptsize 57}\tabularnewline
\hline
\hline
{\scriptsize Rosenbrock $f_{1}$ } & {\scriptsize 52.00\%} & {\scriptsize 0.09} & {\scriptsize 509} & {\scriptsize 861 }\tabularnewline
\hline
\hline
{\scriptsize Rastrigin $f_{2}$} & {\scriptsize 95.00\%} & {\scriptsize 0.01} & {\scriptsize 60} & {\scriptsize 60}\tabularnewline
\hline
\hline
{\scriptsize Griewack $f_{3}$ } & {\scriptsize 100.00\%} & {\scriptsize 0.00} & {\scriptsize 2} & {\scriptsize 2}\tabularnewline
\hline
\hline
{\scriptsize Schaffer $f_{6}$ } & {\scriptsize 100.00\%} & {\scriptsize 0.00} & {\scriptsize 4} & {\scriptsize 4}\tabularnewline
\hline
\hline
{\scriptsize Total} & {\scriptsize 88.50\%} & {\scriptsize 0.02} & {\scriptsize{} 42} & {\scriptsize{} 197}\tabularnewline
\hline
\end{tabular}\label{Flo:goals-best-Rc}\vspace{0.1cm}\caption{Percentage of goals and mean best global values. With the only exception
of $f_{1}$ the mean best global values are successful reaching the
goal $0.01$.}
\end{centering}
\end{table}

\begin{figure}
{\centering
\includegraphics[width=.85\textwidth]{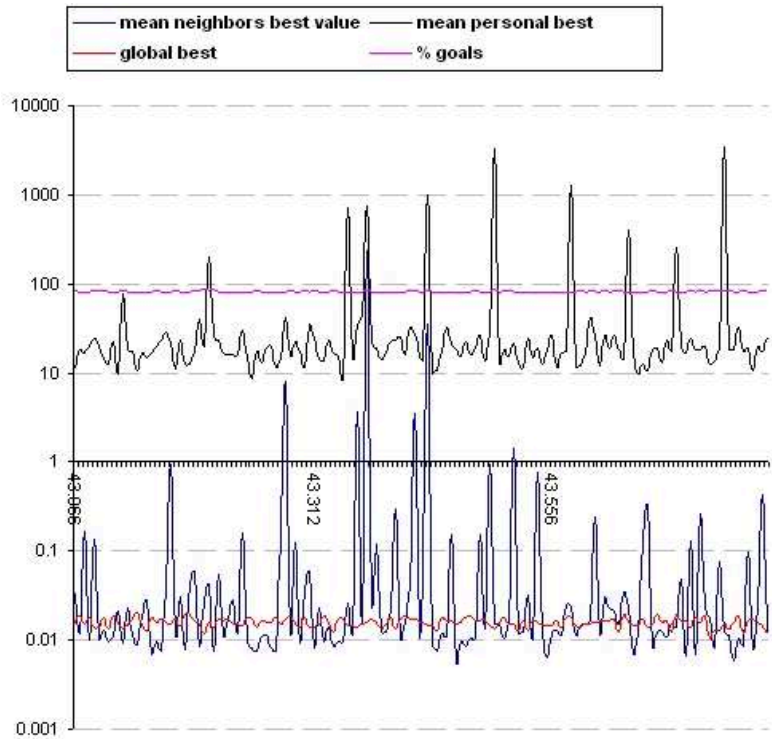}\label{comparison-Rc}
\caption{Values just before $R_{c}\simeq43,758$: Social and personal information
show a great variability, while goals and global best vary slightly. }
}
\end{figure}

In closing, the curves of mean social and personal values are represented
in Fig. \ref{comparison-Rc} together with the global best value
and the percentage of goals. Pearson correlations were calculated
between all the series being the most significant the negative correlation
between goals and global best, -0.385. We conjecture that the lack
of correlation between variables is a characteristic of the interval
around the critical radius.

\section{Exploring R in the the interval $[0,0.9]$ \label{sec:Exploring-R-in}}

As it has been reported in Fig. \ref{Flo:mean-global-all} the minimum value of global best
occurs at $a=0.1,\, R(a)=6.515$. For the same radius, the Fig. \ref{Flo:succesful-runs}
shows a maximum of successful runs. This suggest
that PSO does better with less, but some, social contact: degree $a=0.1$
means that only one in ten individuals has one neighbor at initial
step. The interval $R\in\left[0,\,0.9\right]$ , or equivalently $a\in[0,\:0.001909]$ was
explored to investigate the pass of having no neighbors to have a
minimum of social contact. A total of 91 instances of $R$ with 20
runs per function were executed. Table \ref{table:R_near_0} shows
surprising results: the 94.13\% of the runs were successful in the
interval $\left[0.75,\,0.9\right]$ with the only handicap of longer
iterations, reaching a mean of 258 steps.

\begin{table}
\begin{centering}
\begin{tabular}{|l||c|c|c||c||c|}
\hline
{\scriptsize function} & \multicolumn{4}{c||}{{\scriptsize Goals}} & {\scriptsize Steps all}\tabularnewline
\cline{2-6}
 & {\scriptsize 0.0-0.25} & {\scriptsize 0.26-0.50} & {\scriptsize 0.51-0.75} & \textbf{\em{\scriptsize 0.75-0.90}} & \textbf{\scriptsize 0.75-0.90 }\tabularnewline
\hline
\hline
{\scriptsize Spherical $f_{0}$} & {\scriptsize 60.58\% } & {\scriptsize 91.20\% } & {\scriptsize 99.60\% } & \textbf{\em{\scriptsize 100.00\% }} & \textbf{\scriptsize 262 }\tabularnewline
\hline
{\scriptsize Rosenbrock $f_{1}$ } & {\scriptsize 2.88\% } & {\scriptsize 19.80\% } & {\scriptsize 50.60\% } & \textbf{\em{\scriptsize 70.67\% }} & \textbf{\scriptsize 750}\tabularnewline
\hline
{\scriptsize Rastrigin $f_{2}$} & {\scriptsize 60.19\% } & {\scriptsize 90.00\% } & {\scriptsize 99.80\% } & \textbf{\em{\scriptsize 100.00\% }} & \textbf{\scriptsize 265}\tabularnewline
\hline
{\scriptsize Griewank $f_{3}$ } & {\scriptsize 100.00\% } & {\scriptsize 100.00\% } & {\scriptsize 100.00\% } & \textbf{\em{\scriptsize 100.00\% }} & \textbf{\scriptsize 3}\tabularnewline
\hline
{\scriptsize Schaffer $f_{6}$ } & {\scriptsize 100.00\% } & {\scriptsize 100.00\% } & {\scriptsize 100.00\% } & \textbf{\em{\scriptsize 100.00\% }} & \textbf{\scriptsize 9}\tabularnewline
\hline
{\scriptsize Total} & {\scriptsize 64.73\% } & {\scriptsize 80.20\% } & {\scriptsize 90.00\% } & \textbf{\em{\scriptsize 94.13\% }} & \textbf{\scriptsize 258}\tabularnewline
\hline
\end{tabular}\label{table:R_near_0}\vspace{0.1cm}

\caption{Exploration of $R\in\left[0,\,0.9\right]$ subdivided in intervals: }
\end{centering}
\end{table}

\begin{figure}
\includegraphics[width=1\textwidth]{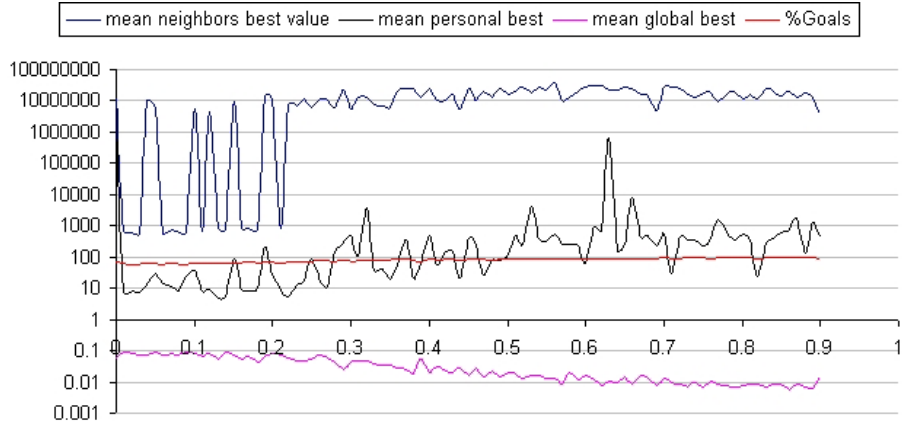}\label{comparison_bests}\caption{Comparison of mean goals and the mean social, personal and global
best for small values of $R\in\left[0,\,0.9\right]$. The minimal
contact with neighbors, for radius over $R=0.6$ ensures the convergence.}
\end{figure}

Figure \ref{comparison_bests} shows that high values of social
information are related to increasing goals and decreasing global
values. To corroborate this tendency, the Pearson coefficient of correlation
was calculated in table \ref{table:pearson}. A very significant positive
correlation exists between the percentage of goals with $R$ and the
mean social best: the major the dispersion of neighbors the major
the goals. Conversely, the goals go in a very significant negative
correlation with global best: the less the best global value the more
the goals . The mean best global value decreases as radius $R$ and
social best increase.

\begin{table}
\begin{centering}
 \begin{footnotesize}\begin{tabular}{|c|c|c|c|c|c|c|}  \hline Correlation & R & Social best & Personal best & Global best & Goals & Steps \tabularnewline
\hline R & 1 & - & - &  - &  - &  - \tabularnewline   \hline Social best & 0.591 & 1 &  - &  - & - & - \tabularnewline  \hline Personal best & -0.122 & 0.024 & 1 & - &  - &-  \tabularnewline   \hline Global best & -0.908 & -0.712 & 0.049  & 1 & - & -  \tabularnewline   \hline Goals & 0.956 & 0.720 & -0.058 & -0.947 & 1 & -   \tabularnewline  \hline Steps & -0.96 & -0.668 & 0.042 & 0.915 & -0.967 & 1 \tabularnewline \hline
\end{tabular}\end{footnotesize}\vspace{0.1cm}

\caption{Pearson coefficients of correlation for dataset $R\in\left[0,\,0.9\right]$.}
\label{table:pearson}
\end{centering}
\end{table}

\section{Comparing different domain through degree $a$.}

Experiments in previous sections were made over the same domain conditions
of $f_{0}$ and $f_{6}$: $x_{max}=100,\, S=200$. Disk percolation
theory gives us a way to relate the expected number of neighbors $a$
to neighborhood radius $R$ of individuals in spite of different values
of parameters of the world, population $P$ and square side $S$,
were supplied. 
In table \ref{table:Rvalues} we report values of $R$
for different degree values $a_1=0.00151$, $a_2=0.8$, $a_3=29$ and at a critical radius $a_{c}=4.512$,
calculated according to  (\ref{eq:area}). Table \ref{table:Rvalues}  shows that conditions $x_{max}=100=\frac{S}{2}$
and $P=30$ give for each $a$ a corresponding radius $R\left(a\right)$
in the intervals studied in previous sections: $R(0.00151)=0.8\in[0.75,\,0.9]$
analyzed in section \ref{sec:Exploring-R-in}, $R(0.8)=18.426$ in
dataset A and $R(29)=110.941$ in dataset B of section \ref{sec:Analysis-of-goals}
and the critical radius $R_{c}=R\left(a_{c}\right)$ in section \ref{sec:Exploring-the-disk}.
We wonder if a change of parameters values will report different patterns
of PSO convergence.

\begin{table}
\begin{centering}
\begin{tabular}{cc}
\begin{footnotesize}
\begin{tabular}{|c|c|c|c|c|}
\hline $f$ & $a_c$ &  $S$ &  $P$ &  $R_c$  \tabularnewline
\hline $f_0$ & 0.00151 & 200 & 30 & 0.801 \tabularnewline
\hline $f_1$ &0.00151 & 60 & 30 & 0.240 \tabularnewline
\hline $f_2$ & 0.00151 & 10.24 & 30 & 0.041 \tabularnewline
\hline $f_3$ & 0.00151 & 1200 & 30 & 4.803 \tabularnewline
\hline $f_6$ & 0.00151 & 200 & 30 & 0.801 \tabularnewline
\hline $f_0$ & 4.512 &  200 &  30 &  43.760 \tabularnewline
\hline  $f_1$ & 4.512 &  60 &  30 &  13.128 \tabularnewline
\hline  $f_2$ & 4.512 &  10.24 &  30 &  2.241 \tabularnewline
\hline  $f_3$ & 4.512 &  1200 &  30 &  262.561 \tabularnewline
\hline  $f_6$ & 4.512 &  200 &  30 &  43.760  \tabularnewline
\hline
\end{tabular}
\end{footnotesize}
 & \begin{footnotesize}
\begin{tabular}{|c|c|c|c|c|}
\hline $f$ & $a_c$ &  $S$ &  $P$ &  $R_c$  \tabularnewline
\hline $f_0$ & 0.800 &  200 &  30 &  18.426 \tabularnewline
\hline  $f_1$ & 0.800 &  60 &  30 &  5.528 \tabularnewline
\hline $f_2$ & 0.800 &  10.24 &  30 &  0.943 \tabularnewline
\hline  $f_3$ & 0.800 &  1200 &  30 &  110.558 \tabularnewline
\hline  $f_6$ & 0.800 &  200 &  30 &  18.426  \tabularnewline
\hline $f_0$ & 29 & 200 & 30 & 110.941\tabularnewline
\hline $f_1$ & 29 & 60 & 30 & 33.282 \tabularnewline
\hline $f_2$ & 29 & 10.24 & 30 & 5.680 \tabularnewline
\hline $f_3$ & 29 & 1200 & 30 & 665.648 \tabularnewline
\hline  $f_6$ & 29 &  200 &  30 &  110.941   \tabularnewline
\hline
\end{tabular}
\end{footnotesize}\tabularnewline
\end{tabular}\vspace{0.1cm}

\caption{Radius values computed for each of the benchmark functions for different values of $a$.}
\label{table:Rvalues}
\end{centering}
\end{table}

\begin{figure}
\includegraphics[width=1\textwidth]{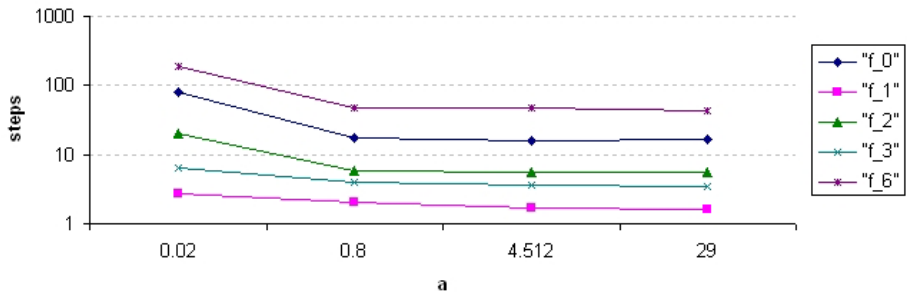}\label{fig:compareTrelea}\caption{Mean steps with Trelea's parameters $w=0.6,\; c_{1}=c_{2}=0.7$.}

\end{figure}

For each value of parameter $a$ the experiment was repeated 500 times,
100 for each function $f_{i}$. All the runs converged with the goal
precision given in \cite{Eberhart200084,Trelea2003317}. Hence,
the only variable to compare is the mean number of steps, given in
Fig. \ref{fig:compareTrelea}. The same pattern of convergence appears:
at small values of $a<R(a)\in[0.75,\:0.9]$ all the experiments converge,
but more steps are needed. From the percolation threshold, a universal
measure, the number of steps slightly decreases.

\section{Conclusions}

We show in this work that percolation threshold is a tool to analyze the convergence of swarms
as a function of the expected number of neighbors at initial step.
While traditionally PSO algorithm is based on the external calculus
of the global best goal by reasons of convergence, we present here
experiments suggesting that, for one hand, very small amounts of information, translated to very low values of
$a$ are enough to provoke the convergence of PSO swarms. Moreover,
better scores of convergence are found. On the other hand, a low degree $a_{c}=4.512$ is enough to both purposes: convergence
to the minimum and short runs. The steps needed for convergence become
constant from this critical value. Hence, more neighbors do the same
than $a_{c}$ neighbors.
Disk percolation threshold is a measure for 2D systems. Efforts will
be made in investigating the calculus of percolation thresholds for
higher dimension. An idea to do that is to compare the area of the
circle of radius $R$ to the surface of the square $\pi R^{2}/S^{2}$
looking for the value of radius $R'$ such that a D-dimensional hypersphere
of this radius maintain the same proportion with respect to the volume
of the hyperboard.

This work can be extended in a further development, comparing previous hybrid PSO approaches and using more benchmark functions. Moreover, we think it is of great interest to study the implications of percolation threshold in fuzzy neurocomputing areas \cite{DBLP:journals/ijon/PedryczA09}, since percolation is considered as a second order state transition phenomenon, having hence a crisp nature.

\bibliographystyle{splncs03}
\bibliography{references}

\begin{thebibliography}{10}
\providecommand{\url}[1]{\texttt{#1}}
\providecommand{\urlprefix}{URL }

\bibitem{bollobs_percolation_2006}
Bollobás, B., Riordan, O.: Percolation. Cambridge University Press, {UK} (2006)

\bibitem{Clerc200258}
Clerc, M., Kennedy, J.: The particle swarm explosion, stability, and
  convergence in a multidimensional complex space. IEEE Transactions on
  Evolutionary Computation  6(1),  58--73 (2002)

\bibitem{DBLP:journals/isci/CorchadoAC10}
Corchado, E., Abraham, A., Ponce Leon Ferreira~de Carvalho, A.C.: Hybrid
  intelligent algorithms and applications. Inf. Sci.  180(14),  2633--2634
  (2010)

\bibitem{DBLP:journals/ijon/CorchadoGW12}
Corchado, E., Gra{\~n}a, M., Wozniak, M.: Editorial: New trends and
  applications on hybrid artificial intelligence systems. Neurocomputing
  75(1),  61--63 (2012)

\bibitem{Eberhart200084}
Eberhart, R.C., Shi, Y.: Comparing inertia weights and constriction factors in
  particle swarm optimization. In: Proceedings of the IEEE Conference on
  Evolutionary Computation, ICEC. vol.~1, pp. 84--88 (2000)

\bibitem{Eberhart200181}
Eberhart, R.C., Shi, Y.: Particle swarm optimization: Developments,
  applications and resources. In: Proceedings of the IEEE Conference on
  Evolutionary Computation, ICEC. vol.~1, pp. 81--86 (2001),
  \url{http://www.scopus.com/inward/record.url?eid=2-s2.0-0034863568&partnerID=40&md5=6ffd3565ce1de09c397454e40a9b315c},
  cited By (since 1996) 890

\bibitem{García_Fernández_Luengo_Herrera_2010}
García, S., Fernández, A., Luengo, J., Herrera, F.: Advanced nonparametric
  tests for multiple comparisons in the design of experiments in computational
  intelligence and data mining: Experimental analysis of power. Information
  Sciences  180(10),  2044--2064 (2010),
  \url{http://linkinghub.elsevier.com/retrieve/pii/S0020025509005404}

\bibitem{Kennedy1997303}
Kennedy, J.: Particle swarm: Social adaptation of knowledge. In: Proceedings of
  the IEEE Conference on Evolutionary Computation, ICEC. pp. 303--308 (1997),
  \url{http://www.scopus.com/inward/record.url?eid=2-s2.0-0030645460&partnerID=40&md5=470218cf1b1e57d9006a17dad075d2ba},
  cited By (since 1996) 436

\bibitem{DBLP:journals/ijon/PedryczA09}
Pedrycz, W., Aliev, R.A.: Logic-oriented neural networks for fuzzy
  neurocomputing. Neurocomputing  73(1-3),  10--23 (2009)

\bibitem{Sahimi_1994}
Sahimi, M.: Applications of Percolation Theory. Taylor and Francis, {UK} (1994)

\bibitem{Stauffer_1992}
Stauffer, D., Aharony, A.: Introduction to Percolation Theory. Taylor and
  Francis, {UK} (1992)

\bibitem{Toscano-Pulido2011179}
Toscano-Pulido, G., Reyes-Medina, A.J., Ram{\'i}rez-Torres, J.G.: A statistical
  study of the effects of neighborhood topologies in particle swarm
  optimization. Studies in Computational Intelligence  343,  179--192 (2011),
  \url{http://www.scopus.com/inward/record.url?eid=2-s2.0-79954459982&partnerID=40&md5=f86b8162a6a53632ffe2abf3537159f0},
  cited By (since 1996) 0

\bibitem{Trelea2003317}
Trelea, I.C.: The particle swarm optimization algorithm: Convergence analysis
  and parameter selection. Information Processing Letters  85(6),  317--325
  (2003),
  \url{http://www.scopus.com/inward/record.url?eid=2-s2.0-0037475094&partnerID=40&md5=9a90a1d64dd52d02eb8c70df5cc5bb88},
  cited By (since 1996) 552

\bibitem{Zhang2010661}
Zhang, L., Mu, H.P., Jiao, C.Y.: Particle swarm optimization with
  highly-clustered scale-free neighborhood model. In: Proceedings - 2010 3rd
  IEEE International Conference on Computer Science and Information Technology,
  ICCSIT 2010. vol.~2, pp. 661--663 (2010),
  \url{http://www.scopus.com/inward/record.url?eid=2-s2.0-77958576686&partnerID=40&md5=c1905cef787722044041a996e207fc61},
  cited By (since 1996) 0

\end{thebibliography}

\end{document}